\documentclass[letterpaper, 10 pt, conference]{ieeeconf}  

\IEEEoverridecommandlockouts                              

\usepackage{comment}
\usepackage{dirtytalk}
\usepackage{graphicx}
\usepackage{makecell}
\title{\LARGE \bf
Express Yourself: Enabling large-scale
public events involving
multi-human-swarm interaction for
social applications with MOSAIX
}

\author{Merihan Alhafnawi$^{1}$, Maca Gomez-Gutierrez$^{2}$, Edmund R. Hunt$^{3}$, Severin Lemaignan$^{4}$, \\ Paul O'Dowd$^{3}$ and Sabine Hauert$^{3}$ 
\thanks{$^{1}$Merihan Alhafnawi is with the Department of Mechanical and Aerospace Engineering,
        Princeton University, Princeton, NJ 08544, USA
        {\tt\small m.alhafnawi@princeton.edu}}%
\thanks{$^{2}$Maca Gomez-Gutierrez is with We The Curious, Millennium Square, Anchor Rd, Bristol BS1 5DB, United Kingdom 
        {\tt\small maca.gomezgutierrez@gmail.com}}%
\thanks{$^{3}$Edmund R. Hunt, Paul O'Dowd and Sabine Hauert are with the Faculty of Engineering, University of Bristol, BS8 1TR, UK 
        {\tt\small \{edmund.hunt,paul.odowd,sabine.hauert\}}
        {\tt\small @bristol.ac.uk}
        }%
\thanks{$^{4}$Severin Lemaignan is with PAL Robotics, Barcelona, Spain
        {\tt\small severin.lemaignan@pal-robotics.com}}%
}

\begin{document}

\maketitle
\thispagestyle{empty}
\pagestyle{empty}

\begin{abstract}

Robot swarms have the potential to help groups of people with social tasks, given their ability to scale to large numbers of robots and users. Developing multi-human-swarm interaction is therefore crucial to support multiple people interacting with the swarm simultaneously - which is an area that is scarcely researched, unlike single-human, single-robot or single-human, multi-robot interaction. Moreover, most robots are still confined to laboratory settings. In this paper, we present our work with MOSAIX, a swarm of robot Tiles, that facilitated ideation at a science museum. 63 robots were used as a swarm of smart sticky notes, collecting input from the public and aggregating it based on themes, providing an evolving visualization tool that engaged visitors and fostered their participation. Our contribution lies in creating a large-scale (63 robots and 294 attendees) public event, with a completely decentralized swarm system in real-life settings. We also discuss learnings we obtained that might help future researchers create multi-human-swarm interaction with the public.

\end{abstract}


\section{INTRODUCTION}

Humans, for centuries, have developed many ways to express themselves to one another to achieve human-human interaction. Such methods include the spoken word, paintings, poetry, acting, and other media. Yet robots could be used as a novel, exciting medium by which humans can express themselves to and through the robots. In group interactions, robot swarms in particular can be helpful, given the swarm's ability to scale to many robots and many users as needed. Social group tasks that involve humans expressing themselves, and which robots could help facilitate, can include group decision-making, brainstorming, collective art-making and education - activities where humans present and discuss their ideas, opinions, and thoughts \cite{mosaix, deliberative_democracy}. 

As such, multi-human-swarm interaction needs to be developed in order to support and facilitate these tasks. While single-human, single-robot interaction in the social robotics field is being extensively studied \cite{socialrobots}, multi-human, multi-robot interaction, (specifically multi-human-swarm interaction), has been scarcely researched. Moreover, swarm robotics research is still mostly confined to laboratory settings and has not been widely tested in real-life \cite{debie2023}. This includes human-swarm interaction, which has been greatly concerned with single-human-swarm interaction so far in laboratory settings or simulation. Furthermore, controlling swarm systems in general is challenging due to their distributed nature \cite{Bashyal08}. Yet achieving such control, without compromising their distributed nature, could lead to useful real-world systems that are robust and scalable to many users and robots. 

To this end, we created MOSAIX \cite{mosaix}, a versatile swarm robotic system that can be used in various social applications by groups of people. MOSAIX is made of 4-inch touchscreen-on-wheels robots, called Tiles. Each Tile has proximity sensors and a camera. We previously used MOSAIX with the public to help users express their opinions, create collective art and for education \cite{mosaix}. 

In this paper, we present our latest work with MOSAIX in collaboration with We The Curious, a science museum in Bristol, UK. The museum sought an innovative way to engage visitors while collecting their insight on how the city could improve residents’ health and well-being. Those insights could serve as ideas for future exhibits for the museum. MOSAIX was a perfect fit, as it can both collect input and visually represent this input to the public, thereby increasing engagement and fostering meaningful participation. We used the Tiles of MOSAIX as smart sticky notes, enabling users to enter their ideas. The robots then autonomously aggregated based on different themes, creating a dynamic and evolving idea pool to engage visitors throughout the day. We successfully deployed 63 Tiles in the museum over 3 days, collecting around 315 ideas from approximately 294 attendees. This application of MOSAIX could be utilized in the future by teams, for example in corporations, to help with the ideation process of brainstorming.

Our contribution, lies in developing the first -to our knowledge- completely decentralized, large-scale (63 robots), multi-human-swarm interaction with members of the public in the real world. We further present lessons learned from our accumulated experience from this public event, as well as previous ones presented in \cite{mosaix}. These insights are intended to assist future researchers in developing smooth large-scale public events with swarm robots.

\section{Related Work} \label{sec:related_work}

Human-swarm interaction is an emerging field that has been increasingly studied in the past few years \cite{kolling16}. Platforms were created that enable human-swarm or human-multi-robot interactions in social applications. For example, tangible interfaces, such as \say{Zooids} \cite{zooids16} and \say{ShapeBots} \cite{shapebots}, help people in their daily lives. While these 2 platforms share with our work the theme of facilitating tasks for people, they have a centralized control. Our work differs in using a decentralized control, avoiding central points of failure. 

\say{Thymio} is another multi-robot platform used for education \cite{thymio}. Thymios are also centrally controlled. However, Thymios were later used as a swarm system by exploiting the IR proximity sensors to send small payloads of information \cite{vitanza19}. A swarm behaviour was also implemented with 8 Thymios by demonstrating the collective decision-making of honeybees \cite{seely12}. With our work, we employ robots on a large scale (63 robots). There are other multi-robot systems built for educational applications, such as \say{microMVP} \cite{micromvp} and \say{Cellulo} \cite{cellulo17}. However, these systems adopt a centralized control. 

Glowbots, created by \cite{glowBots2008}, is another system that consist of see-Pucks robots \cite{seePucks2008} which are e-Pucks \cite{epuck} that have 148 LEDs to show patterns. A human can interact with the robots by shaking individual robots to either influence the swarm with the robot or get influenced by the swarm. Robotic Canvas \cite{robotic_canvas} helps users create paintings with Kilobots \cite{kilobots2012} by using hand gestures and dedicated robots. In later work, the system could also detect and represent projected shapes \cite{saliency}. Another swarm system was designed to assist humans in creating paintings by using robots equipped with painting materials \cite{santos2}. Like MOSAIX, these interactions are physical, proximal and decentralized. However, MOSAIX's touchscreens and sensors allow for creating intuitive modes of interaction (such as entering input through a virtual keyboard).

Few systems have explored multi-human, multi-robot interaction. For example, \cite{multi1} and \cite{multi2} implemented a system where multiple users can use multiple robots to transfer objects. The user studies with said system focused more on analysing inter-human dynamics of multi-human, multi-robot systems, such as whether humans prefer to work freely, or be assigned roles \cite{multi1}, and the kind of communication the human preferred (directly communicating with one another or communicating through the interface) \cite{multi2}. Work in \cite{lewis10} also created multi-human, multi-robot interaction to understand how humans preferred organising their teams. Likewise, work in \cite{miyauchi2023} explored sharing robots from a simulated swarm among 2 participants. Another system in simulation explored the coordination between human teams, ground robots and UAVs in a military application \cite{freedy08}. We are instead interested in integrating robots into the existing environment of humans to help them with social tasks from their daily lives, by facilitating these tasks. We are also interested in directly deploying these robots into the real world rather than in simulation or laboratory environments.

\section{Engaging with the public}

We The Curious, a science museum at Bristol, UK, aimed to collect public input in an engaging way to encourage participation on how the city could improve residents' health. The public's opinion could help the museum shape future exhibits. We asked the public \say{What can the city of Bristol do to provide a healthier lifestyle for its citizens?}. The Tiles of MOSAIX were used as interactive smart sticky notes to capture ideas and autonomously create aggregates of similar themes. 

\begin{figure*}
   \centering
          \includegraphics[scale=0.45]{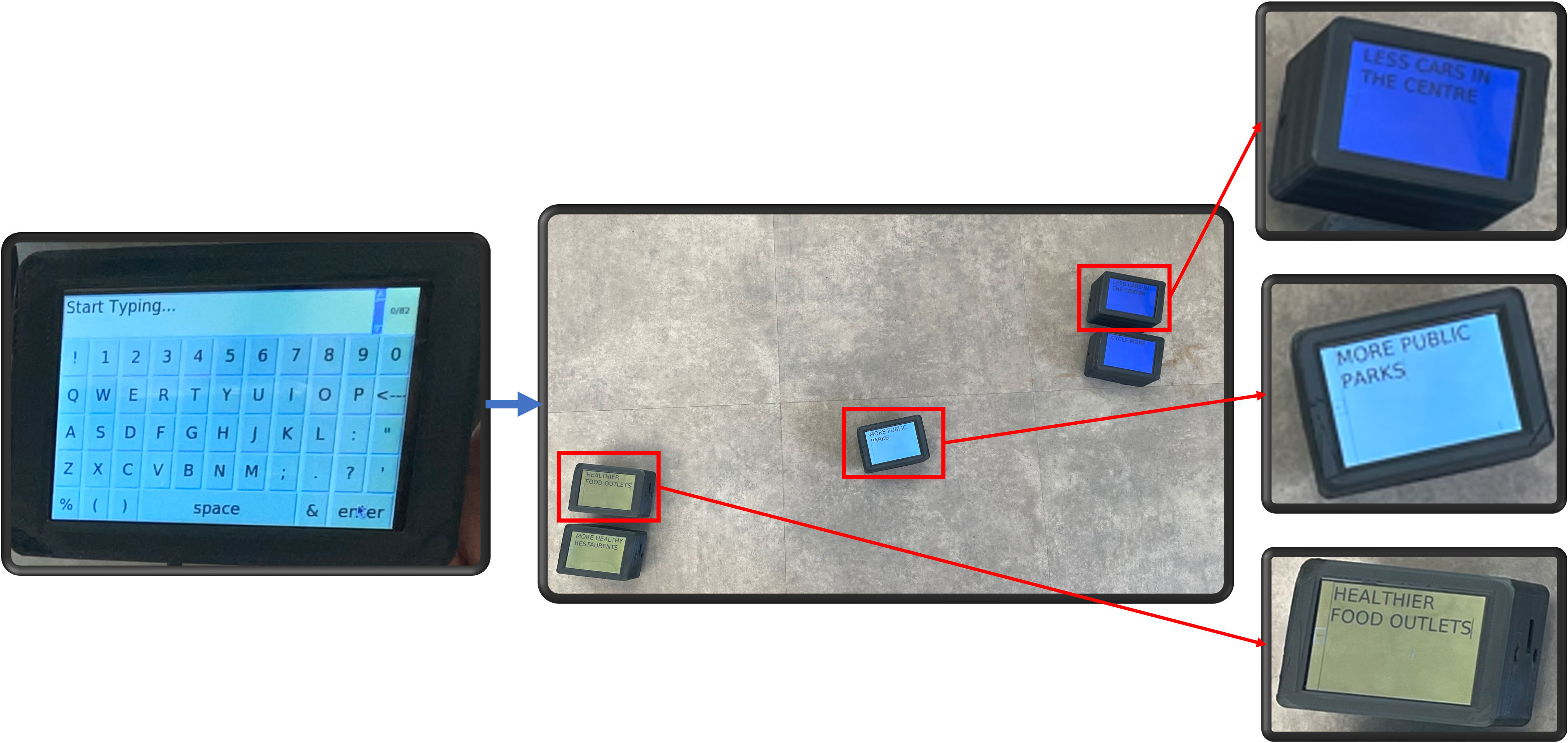}
          \caption{Left: Virtual keyboard shown to participants to enter their idea. Middle: Aggregates formed by robots from participants' ideas. Right: samples of ideas shown. The small button on the left of the screen allows participants to go back to the virtual keyboard and re-enter ideas.}
          \label{fig:keyboard}
\end{figure*}

Participants were handed a Tile that showed a virtual keyboard (see Fig. \ref{fig:keyboard}) to enter their idea onto. Once they entered their idea and then pressed \say{enter}, their idea appeared on full-screen, next to a button that allowed them to go back to the virtual keyboard and re-enter an idea if they wish. They then put the robot down on the floor in the middle of the arena, and the robot rotated to look for the aggregate it needed to join, and autonomously moved to the aggregate. We then requested that participants write the same idea on a traditional sticky note and place it on a board, aligning it with related ideas themselves. Finally, we asked the participants to take a short questionnaire (shown in Table \ref{tab:brain_question}) to compare their experience with smart vs traditional sticky notes. We obtained ethical approval by the University of Bristol's Faculty of Engineering Research Ethics Committee (FREC) to run this event.

 \begin{table}
  \vspace{2.5mm}
  \caption{\label{tab:brain_question} Ideation post-experiment questionnaire }
\centering
\begin{tabular}{||l l||}
 \hline
  \thead{Questions}  & \thead{Choices} \\ 
  \hline\hline
   \makecell{1- Robots autonomously organizing \\ themselves into clusters would be helpful \\ in brainstorming sessions. }  & \makecell{Completely Disagree, \\ Disagree,  Undecided, \\ Agree, Completely Agree} \\
  \hline
  \makecell{2- How easy or hard was it to use \\ the robotic system to enter ideas \\ and traverse through them?} & \makecell{Very Hard, Hard, \\ Neutral, Easy, Very Easy} \\
  \hline
   \makecell{3- Which system \\ performed better clusters?} & \makecell{Robotic, Traditional} \\
  \hline
  \makecell{4- Overall, which system \\ did you prefer?}  & \makecell{Robotic, Traditional} \\
   \hline
   \makecell{5- Why did you prefer \\ that specific system?} & \makecell{Open answer} \\
  \hline
  \makecell{6- Any feedback for us?} & \makecell{Open answer} \\   \hline
  \end{tabular}
\end{table}

\subsection{Software Implementation}

The Tiles use an ad-hoc network, where each Tile acts as a node in the network (no router necessary). Each Tile broadcasts the following to the ad-hoc network: its robot ID, aggregate number (starting at 0 if the Tile did not find a suitable aggregate) and the idea entered on it by the participant. This message is encoded into a string format to be broadcast, for which all robots are programmed to decode. 

Natural Language Processing running on each Tile is used to extract sentence embeddings, using SentenceTransformers Python Framework \cite{embeddings}, of an idea when it is entered onto a Tile. Then, cosine similarity is used to measure how similar this idea is to all other ideas (which were broadcast over the network by other Tiles). The result is a similarity score on a range from 0 to 1, where 0 means the ideas are absolutely dissimilar and 1 means the ideas are identical. The Tile then moves to the aggregate of the robot that scored highest in similarity. The highest score has to be higher than 0.3, since while testing the system, we found that if the threshold was lower, many ideas that were dissimilar were deemed similar and vice versa when the threshold was higher. There were 5 AruCo markers (more can be added) that act as rendezvous points for the robots to form aggregates. The robot uses its camera to recognize, and move to, the aggregate by recognizing the AruCo markers associated with that aggregate (see Fig. \ref{fig:alldays} where robots are aggregating by AruCo markers). Every aggregate utilizes a different touchscreen background colour to act as a further identifier of aggregates, in addition to physically aggregating under an AruCo marker. The robot uses its front proximity sensor to avoid obstacles while navigating to the aggregate. 

If a robot cannot find a similar theme or is the first to have an idea, it will seek out an empty AruCo marker —one not yet associated with any robots— and create its own aggregate (i.e., a new theme) that other robots can join. The robot will then change its background colour to one not currently used by any other aggregate. If all aggregates have been taken, the robot does not move, and changes its background colour to white.  It then keeps recalculating similarities from other robots' ideas. This is done in case an aggregate empties out (the robots are physically removed or matches with new aggregates). This allows the system to evolve and be adaptable to change.

\subsection{Results and discussion}
\begin{figure*} 
      \centering
          \includegraphics[scale=0.48]{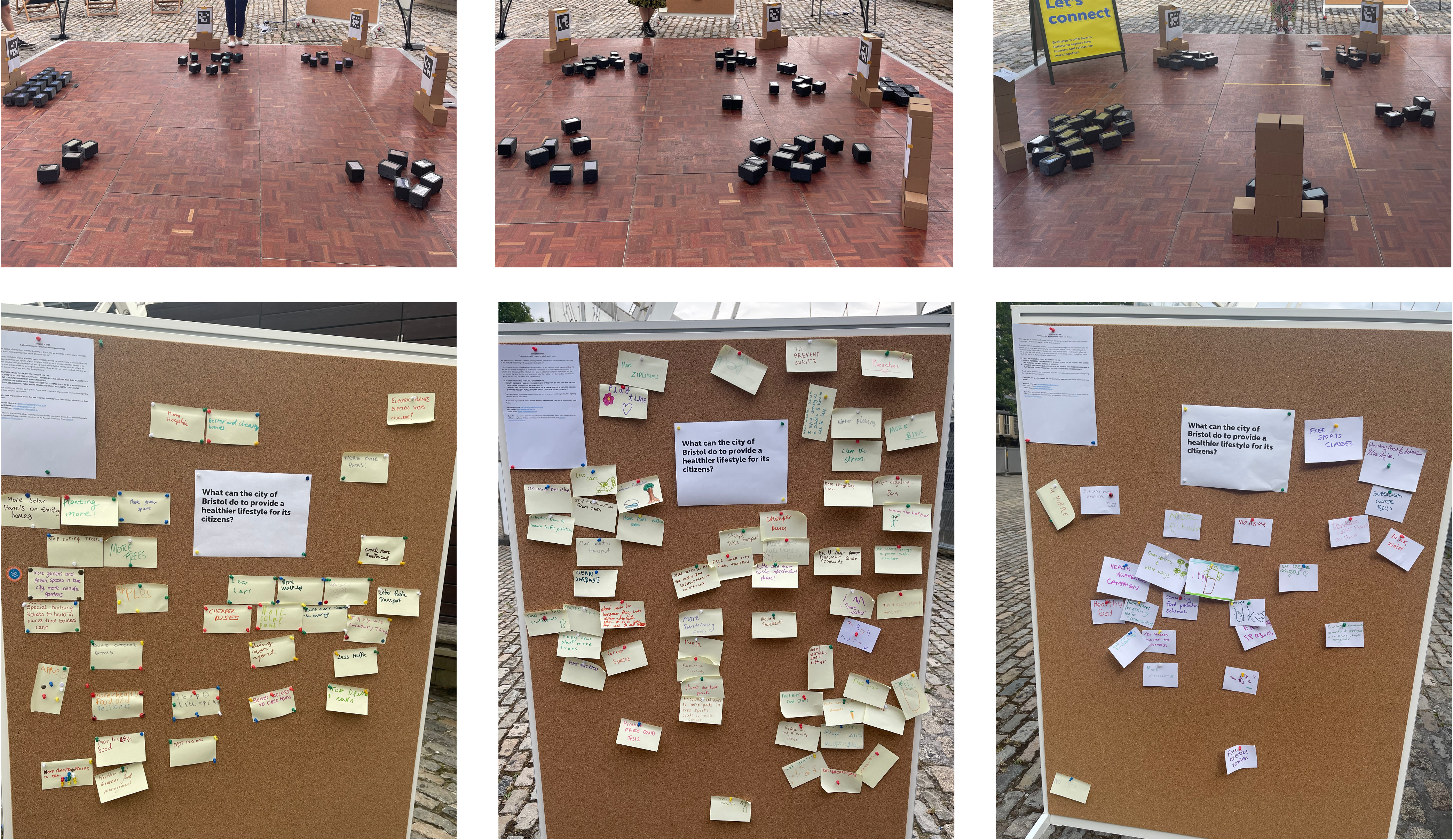}
          \caption{Last aggregates formed at the end of day 1, day 2 and day 3, respectively. Top: Robotic aggregates. Bottom: Traditional aggregates.}
          \label{fig:alldays}
   \end{figure*}

As requested by the museum, we wanted the experience to be engaging to visitors while collecting their input. The event successfully ran for 3 days where we collected around 315 ideas from approximately 294 attendees, of which 50 consented to taking the questionnaire. We ran the event for 4 hours on the first day and for 5 hours on the second and third days; a total of 14 hours. Up to 63 robots were used. Fig. \ref{fig:alldays} shows the robotic and traditional aggregates formed at the end of each day.

\begin{figure}[h]
      \centering
          \includegraphics[scale=0.7]{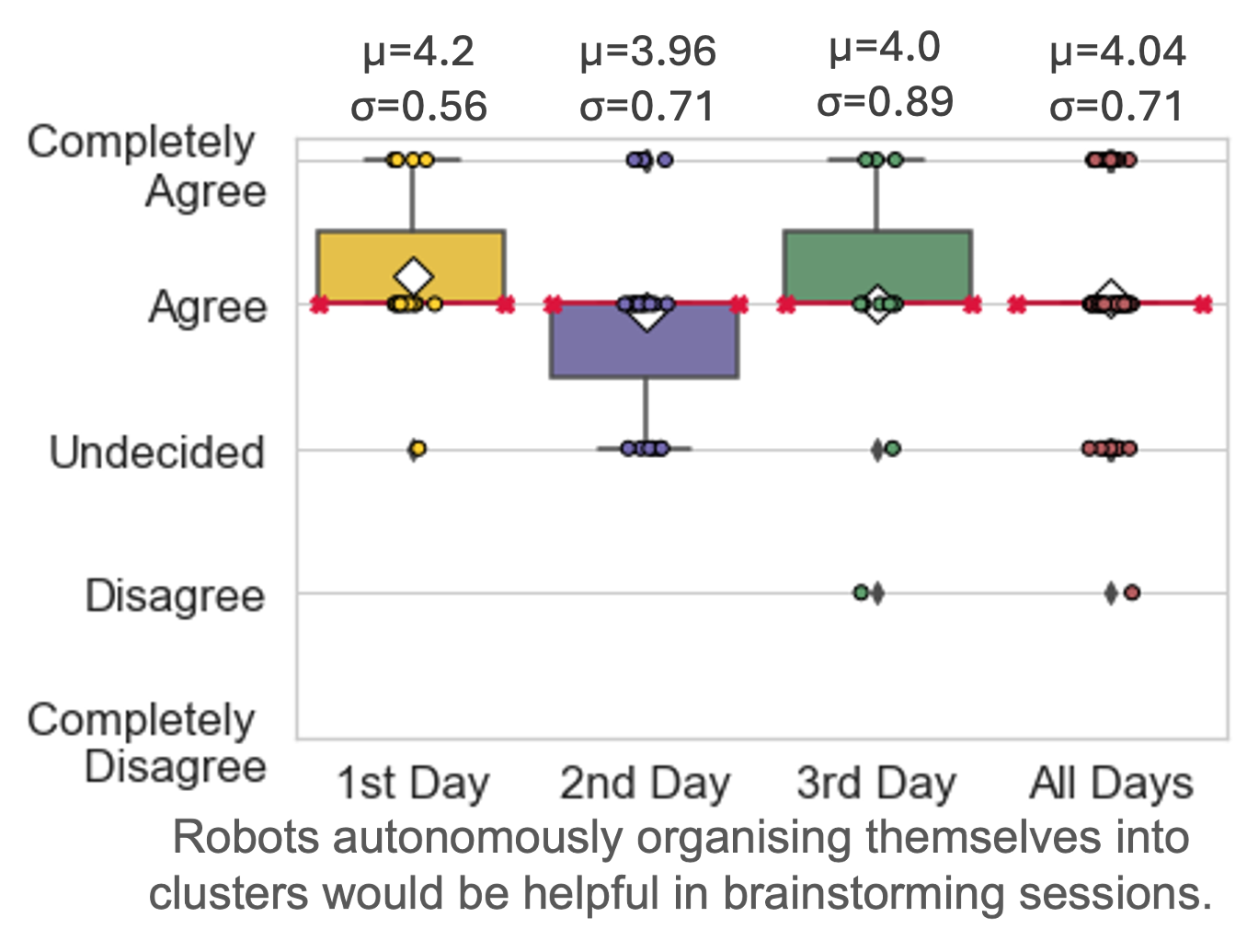}
          
          \caption{ Boxplot figures showing the answers to question 1 of the questionnaire given to 50 participants. The dots represent the participants’ answers, the red line ending with crosses is the median of the participants’ answers, and the white diamond is the mean of the answer.}
          \label{fig:helpful}
\end{figure}

\begin{figure}[h]
      \centering
          \includegraphics[scale=0.7]{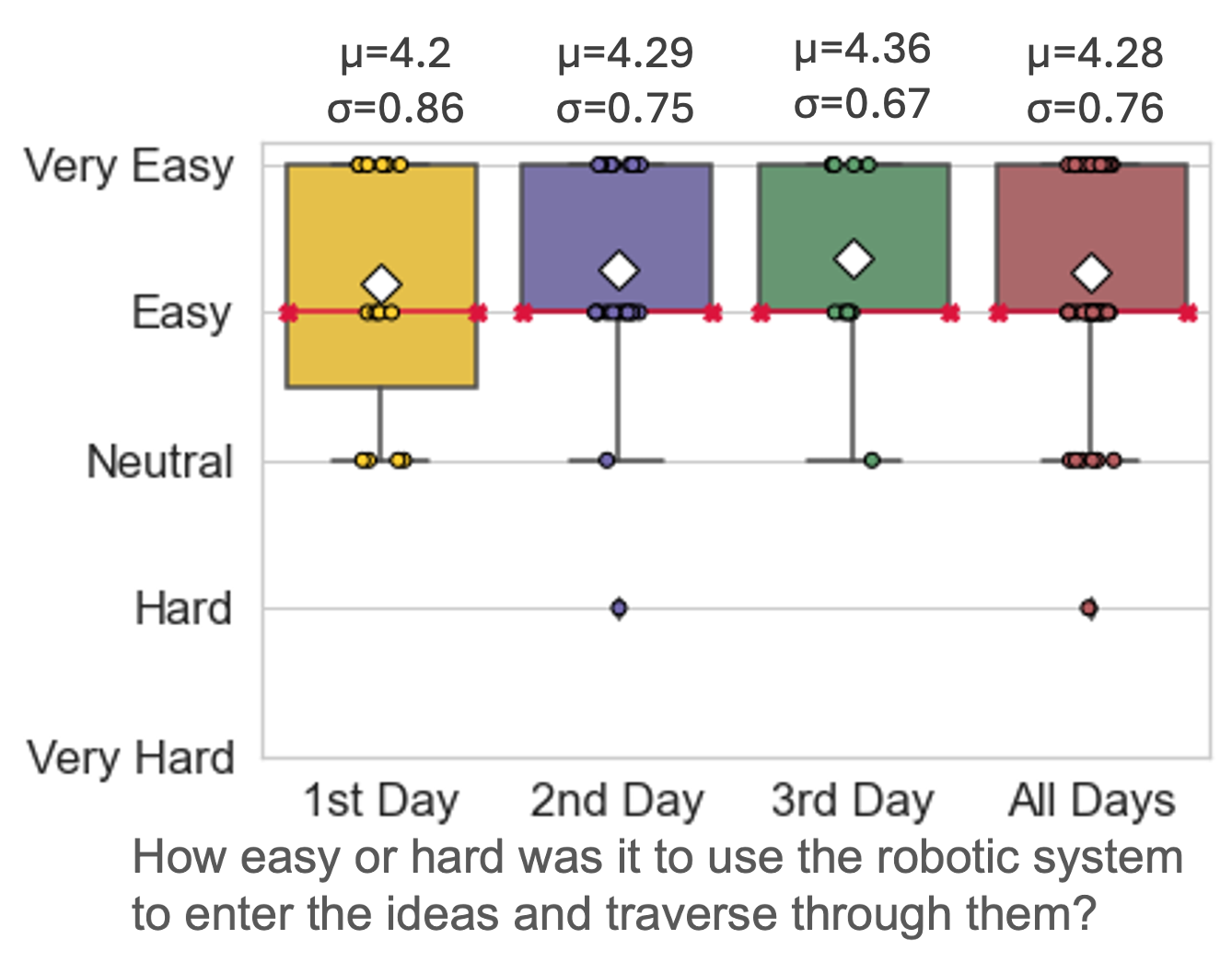}
          \caption{Boxplot figures showing the answers to question 2 of the questionnaire given to 50 participants. }
          \label{fig:easy_hard}
\end{figure}

Fig. \ref{fig:helpful} and Fig. \ref{fig:easy_hard} show the results of the first 2 questions (shown in Table \ref{tab:brain_question}) of the questionnaire. We report and discuss the mean ($\mu$) and standard deviation ($\sigma$) on a scale from 1-5, which correspond to the Likert scale used. For example, 1 represents \say{Completely Disagree} for the first question and \say{Very Hard} for the second question and 5 represents \say{Completely Agree} for the first question and \say{Very Easy} for the second question. We also report and discuss the medians \cite{schrum20}. 

Participants thought the robots were easy to use, as is evident from Fig \ref{fig:easy_hard}, where both the average and the median were at \say{Easy} for all 3 days. One participant commented that they prefer the robotic system because \say{it's more easy to do}. This entails that the robot design did not hinder system usability. This is a key result we wanted to observe with MOSAIX, as we wanted to ensure that the robots will smoothly facilitate multi-human-swarm interaction and not hinder usability. As can be seen from Fig. \ref{fig:helpful}, most people agreed that it would be helpful for robots to arrange themselves autonomously during brainstorming sessions. We also observed sometimes that people returned after a while to see how the aggregates are performing and enter more ideas. This suggest that users were indeed engaged with the system and recognized its potential value in brainstorming.

\begin{figure}[h]
      \centering
          \includegraphics[scale=0.450]{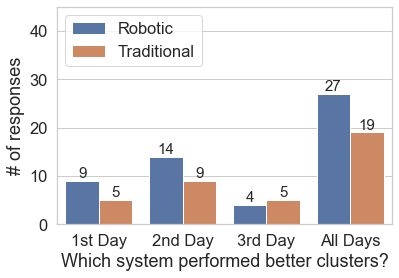}
          \caption{Barchart figures showing the answers to question 3 of the questionnaire given to 50 participants. Note that the total number of response for robotic and traditional combined (n=46) is less than 50. This is because participants were free to skip questions they did not want to answer. }
          \label{fig:better_clusters}
\end{figure}

\begin{figure}[h]
      \centering
          \includegraphics[scale=0.45]{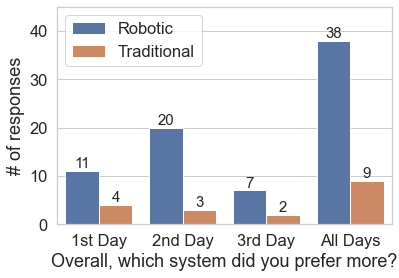}
          \caption{Barchart figures showing the answers to question 4 of the questionnaire given to 50 participants.}
          \label{fig:which_system}
\end{figure}

Fig. \ref{fig:better_clusters} and Fig. \ref{fig:which_system} show the results of the 3rd and 4th question on the questionnaire. As evident from Fig. 6, the robotic system was overall highly preferred over the traditional system, with 81\% choosing the robotic system as their preference. As shown in Fig. 5, more people (57\%) thought that the robots created better clusters (i.e., aggregates) compared to the traditional system. However, the difference in opinions on this question was not as pronounced as on the previous one. Some participants mentioned that \say{[I] Really liked the concept - good to see different links being made but the robots missed some obvious ones}. While the robots performed the path planning motion and physically merging with clusters well, some aggregates might not have merged as expected by human participants. For example, a participant pointed out that \say{it doesn't understand the meaning of my answer (street workout park). [the robot] thought it was more \say{natural} meaning}. In this case, the robot had joined an aggregate with a nature theme since the idea had the word \say{park} in it. Another participant pointed out that the robots missed links between ideas by answering \say{I could place between clusters due to nuance the robots don't have... Yet!} for why they chose the traditional system. The subtle meanings behind words or ideas, or cultural nuances, may not be comprehended by the robots. However, we still had participants that mentioned that the robots \say{found the answers better} and another that mentioned that the robots \say{promoted discussion} and \say{it also forced you to look at neighbouring ideas to see if they fit}. In this regard, the application of a robotic system may still stimulate new associations and facilitate discussions. For future work, other methods, such as Large Language Models, could be tested to analyze the ideas and observe how these methods would affect the participants' opinions on aggregate formation in a dedicated user study.

Regarding why participants preferred the robotic system, responses to question 5 revealed that 18 individuals mentioned the word \say{fun} as a reason for their choice. Notably, 6 of these respondents preferred the robotic system overall but felt that the traditional system performed better clusters. Moreover, a participant mentioned that \say{[the robotic system] is more exciting than traditional methods}, and another participant said \say{Fun to have it automated, stand back and watch}, while another mentioned \say{more impressive}. We believe these comments stem from the fact that it is a new experience for people. 

The system was received well overall by the museum visitors, and made for a compelling and stimulating exhibit. 5 responses mentioned the word \say{engage} or \say{engaging} regarding the robotic system. Also, participants commented on the interactivity  of the system by saying \say{great interactive activity}, \say{More interactive so makes you feel that you are a participant} and \say{Robotic = new- still works like traditional, but with interactive responses}. A participant mentioned \say{novel ideas makes it less tedious}, while another mentioned that the robots \say{match for you}. Other participants were excited to interact with a new technology, mentioning comments like they preferred the robotic system because \say{[robotic system is] more modern}, \say{[robotic system is] exciting, new}, \say{Robotics is the future!}, \say{I'm a robotics enthusiast, loved it!}, \say{Tech aspect and automation}, \say{Because robots are cool}, and \say{Traditional is better in current settings but future is more autonomous more modern}. In addition, participants commented on how fast the robots found and merged with their aggregate, as an answer to why they preferred the robotic system, with comments such as: \say{[robotic system is] much quicker}, \say{easy, fast, entertaining}, and \say{instant decision}.

Therefore, we conclude that members of the public found the system easy to use, and the majority preferred the robotic system (more than 80\%) for reasons such as its novelty, engagement, and interactivity. While some participants thought that the clustering was made better using the traditional system due to the robots missing some obvious links, some of them were excited by the system and saw its potential. Participants were also excited by the use of technology and robotics, and appreciated the system being responsive and quick. We therefore created a tool that, as the museum required, engaged the public while collecting and visualizing their opinions. Such a tool can be enhanced in the future and used in brainstorming sessions to facilitate idea development.

\section{Lessons learned from creating multi-human-swarm interaction with the public} \label{sec:learnings}

 We accumulated some learnings from hosting public engagement events. These gained insights helped us easily and smoothly deploy the robots for both the operators and the users. In this section, we present 4 key learnings. 

\subsection{Decentralization and scalability}

Multi-human-swarm interaction involves processes that could happen simultaneously. These processes could also increase proportionally as more human users and robots are added to the system. Therefore, we found that decentralization helped in preventing bottlenecks in the system that could occur from having a single point of control. Bottlenecks could cause delays in system response which could frustrate users \cite{yang17}. Our system, having each robot process on its own rather than depend on a central controller, provided a fast response that users appreciated.

Experiments done in the real world, especially involving many people at once, can be fast-paced and overwhelming. It is crucial to be able to solve the problem quickly and efficiently, if and when it arises. This is especially true in settings such as museum exhibits where people can just leave if they are too frustrated with the system or waited too long for an issue to be resolved.  With swarm systems, decentralization provides a great advantage as there are no single points of failure and it is easy to replace a robot with another if it fails without having to compromise the entire system, or take too much time to integrate and deploy new robots. Also, swarm scalability helped us increase the number of robots whenever needed as more people showed up to the events to accommodate them based on the number of people and interactions without having to change any parameter of the system or the software program.

However, there might be limitations in the future regarding scalability that we did not face. We had a small arena (since all robots need to be in the same area as the humans) and so we did not face any fragmentation to the ad-hoc network. Therefore, we advice future swarm researchers to take advantage of the distributed nature of the system while also taking into account, and mitigating, problems that could affect their systems with adding a very large number of robots, as also discussed by \cite{Bjerknes2013}. 

\subsection{Novelty}
In experiments carried out by \cite{luria17}, results showed that people found a robot to be the most engaging and had the highest interaction flow, while a touchscreen and a mobile application were the most usable in a smart-home scenario. The authors then suggest there should be a trade off between flow (robots) and usability (touchscreen/mobile applications). Another research by \cite{Zhexenova20} showed that children preferred a robot over a tablet while learning and there was a positive mood change with the robotic interaction. 

 These results align with previous results obtained with MOSAIX where people were engaged more with moving robots over a system that imitated a tablet-like experience \cite{deliberative_democracy}. They also align with the results discussed in the previous section with MOSAIX, where people preferred the robotic system over the traditional system. People also found the robots were easy to use (given intuitive interactions with the touchscreen). Hence, we can infer from results by \cite{luria17}, \cite{Zhexenova20}, and from our own results, that people are likely to be more engaged in interactions that are new to them and use novel aspects.

However, long-term studies might be needed to explore the effects of habituation on users of specific systems, such as studies by \cite{koay07}. Results may show the effect on future engagement with the system if used many times over a long period. Therefore, our advice to human-swarm interaction researchers is to make the interaction intuitive and easy to use, while creating a new experience to humans that will keep them interested and engaged for a long time. Our observation was that people generally are curious and willing to try new technology. The challenge might be to have them adopt such technologies for the long term.

\subsection{Flexibility for users and operators}

A swarm system involves autonomous agents working together to achieve a task \cite{brambilla13}. We believe that with systems such as ours that facilitate social tasks, flexible autonomy, a concept discussed by \cite{hussein18}, could be useful. For example, in the ideation application, humans could physically move a robot to another aggregate, but the robot would not have registered that it now belongs to another aggregate. However, it could be interesting to analyze how the aggregates would have evolved if humans could overwrite the robot's decision of which aggregate it belongs to. It could also act as a point of learning for the system to be able to constantly improve the model from human input. 

Therefore, we advice researchers to explore how swarm systems can learn and adapt to humans if possible. That notion alone might increase adoption and long term use of the system as it learns from humans (avoiding, of course, over-fitting).

\subsection{Easy, simple deployment}

Not only was it crucial to have easy-to-use systems for participants, but it was also important to have an easy-to-deploy system for operators. For example, during the event, we only had to turn on a Tile via its switch, which then showed a screen, that gave the option to operators to either simply start the program or debug the robot. So, to add a robot to the system, one only had to open a switch and press a button. This made the operators, who were not just the researchers but also museum staff, deploy robots quickly and easily to the swarm.

Lastly, one of the important factors that made MOSAIX easy to deploy was its ability to seamlessly integrate in existing environments. There was no need to change the environment itself or for heavy equipment to be deployed. This was how MOSAIX was used at a classroom (education) and at a shopping mall (opinion-mixing) in previous work \cite{mosaix}, and at a science museum.

\section{Conclusion}

As robots become ubiquitous in people's lives in the future, it is important to study and develop the interactions between groups of people and groups of robots. In this work, we deployed a swarm of 63 robots—Tiles from the MOSAIX system—at a science museum for 3 days to engage with the public. These robots functioned as interactive smart sticky notes, collecting and visualizing visitors' input to facilitate the ideation process. By serving as a dynamic visualization tool, the robots kept visitors engaged and encouraged them to share their opinions and also assisted the museum in gathering valuable insights for future exhibits. In total, we collected 315 ideas from 294 attendees throughout the 3 days.

Results showed that over 80\% preferred the robotic sticky notes over traditional ones, and that overall, the visitors found the robots engaging, exciting and easy to use. We had accumulated some valuable insights from deploying the robots in public events. We shared these learnings, hoping they could help future researchers design their own public events with swarm robots.

We aim to continue using MOSAIX as a tool to allow humans and robots to “think together” in a multitude of ways. Therefore, future work could look at how to solicit continued user engagement, such as agitating the swarm to form new aggregates, or requesting aggregates on themes prompted by users.

\bibliographystyle{IEEEtran}
\bibliography{root}

\end{document}